\documentclass[conference]{IEEEtran}
\IEEEoverridecommandlockouts
\usepackage{cite}
\usepackage{amsmath,amssymb,amsfonts}
\usepackage{algorithmic}
\usepackage{graphicx,subfigure}
\usepackage{textcomp}
\usepackage{xcolor}
\usepackage{booktabs} 
\usepackage[normalem]{ulem}
\useunder{\uline}{\ul}{}
\usepackage{multirow}
\usepackage{amssymb}
\usepackage{epsfig}
\usepackage{bbm}
\usepackage{caption}
\usepackage{comment}

\usepackage[normalem]{ulem}

\usepackage{authblk}
\usepackage{floatrow}
\newfloatcommand{capbtabbox}{table}[][\FBwidth]
\def\BibTeX{{\rm B\kern-.05em{\sc i\kern-.025em b}\kern-.08em
    T\kern-.1667em\lower.7ex\hbox{E}\kern-.125emX}}
\begin{document}

\title{CvS: Classification via Segmentation For Small Datasets}

\author[1]{Nooshin Mojab}
\author[1]{Philip S. Yu}
\author[2]{Joelle A. Hallak}
\author[2]{Darvin Yi}
\affil[1]{Department of Computer Science, University of Illinois at Chicago, Chicago, IL, US }
\affil[2]{Department of Ophthalmology and Visual Sciences, University of Illinois at Chicago, Chicago, IL, US  \authorcr Email: {\tt \{nmojab2, psyu, joelle, dyi9\}@uic.edu} \vspace{-2ex}}

\maketitle

\begin{abstract}
Deep learning models have shown promising results in a wide range of computer vision applications across various domains. The success of deep learning methods relies heavily on the availability of a large amount of data. Deep neural networks are prone to overfitting when data is scarce. This problem becomes even more severe for neural network with classification head with access to only a few data points. However, acquiring large-scale datasets is very challenging, laborious, or even infeasible in some domains. Hence, developing classifiers that are able to perform well in small data regimes is crucial for applications with limited data. This paper presents CvS, a cost-effective classifier for small datasets that derives the classification labels from predicting the segmentation maps. We employ the label propagation method to achieve a fully segmented dataset with only a handful of manually segmented data. We evaluate the effectiveness of our framework on diverse problems showing that CvS is able to achieve much higher classification results compared to previous methods when given only a handful of examples.
\end{abstract}


\section{Introduction}
\label{sec:intro}
Over the past decade, deep learning algorithms have been proved to excel at various computer vision tasks ranging from classification to object detection and segmentation. The tremendous success of deep learning algorithms in computer vision has inspired great innovations across many domains from healthcare to the automotive industry. Contemporary deep neural networks heavily rely on a large amount of data to learn robust models that generalize well to unseen data. For this reason, large-scale datasets have been collected \cite{deng2009imagenet,lin2014microsoft, lecun2010mnist, krizhevsky2009learning}, enabling the development of powerful models pushing the state-of-the-art further in many computer vision applications. However, collecting large-scale datasets is not only laborious but also infeasible in settings where the scarcity of data is inevitable due to the nature of the task such as the diagnosis of a rare disease. On the other hand, the deep neural network tends to overfit when data is scarce. This problem can be even more severe for classification networks modeled by deep neural networks. Therefore, developing a classifier network that is able to handle small datasets is crucial for applications with limited data. 

Multi-task learning has shown to be an effective way of exploiting the knowledge in related tasks to improve the generalization of each task by jointly training the network with a group of related tasks \cite{zhang2018overview, zhang2021survey}. Segmentation are among the ones that are commonly being employed with classification in a multi-task setting \cite{mojab2019deep}. Despite the success of MTL methods in some applications, balancing the loss from different heads often lead to overfitting the problem and hence limits their applications. Moreover, acquiring segmented data is not only very time-consuming but also requires the knowledge of domain experts in many fields. 

In general, segmentation models have shown more robustness to overfitting when trained with a small dataset \cite{ronneberger2015u, badrinarayanan2017segnet}. The reason could potentially be attributed to encoding a dense pixel-wise loss that incorporates a high bias shape prior to the learning process. In this paper, we propose a novel framework, called CvS (Classification via Segmentation), that leverages the power of segmentation to solve the classification task in a low data regime. CvS is not only able to outperforms previous works, but it also addresses the problem of loss balancing in MTL networks and the difficulty to procure segmentation labels for a dataset.

\textbf{Main contributions.} In this work, (i) we introduce a novel framework, called CvS, that utilizes the segmentation power to perform classification when the model has access to only a handful of examples ($\sim1-5$ samples per class). (ii) As opposed to the standard MTL framework, CvS is a single-headed approach forcing both tasks, segmentation, and classification, to be computed together. (iii) We employ a label propagation technique to obtain fully segmented data by segmenting only a small subset of the dataset. 

We show the effectiveness of our approach through extensive experiments on a diverse set of problems in the context of image classification.  
 
\section{Related Work}
The machine learning field has witnessed an evolution of deep classifier networks over the past decades evolving from a simple multi-layer network to more complex and deeper neural networks aiming at pushing the state-of-the-art for the image classification problems \cite{lecun1990handwritten,krizhevsky2012imagenet, simonyan2014very, szegedy2015going, szegedy2016rethinking, he2016deep, huang2017densely}. The current state-of-the-art models for image classification, however, are mainly developed based on the assumption of the availability of a large amount of data \cite{pham2020meta, foret2020sharpness, dosovitskiy2020image, byerly2020branching, mazzia2021efficient}. 
\begin{figure*}[ht]
\begin{center}
\subfigure[Classification Via Segmentation (CvS)]{\label{fig:cvs}\includegraphics[width=120mm]{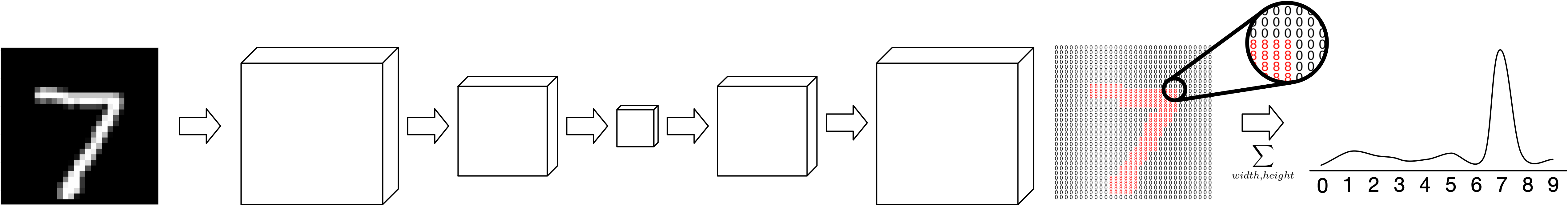}}
\end{center}
\begin{center}
\subfigure[Standard Classification Network ]{\label{fig:classification}\includegraphics[width=90mm]{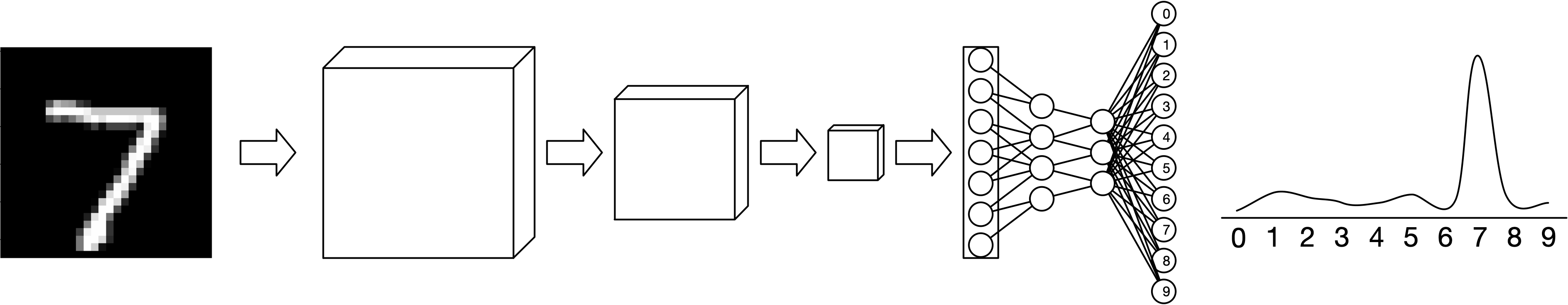}}
\end{center}
\begin{center}
\subfigure[Multi-task Learning Network]{\label{fig:multi-task}\includegraphics[width=90mm]{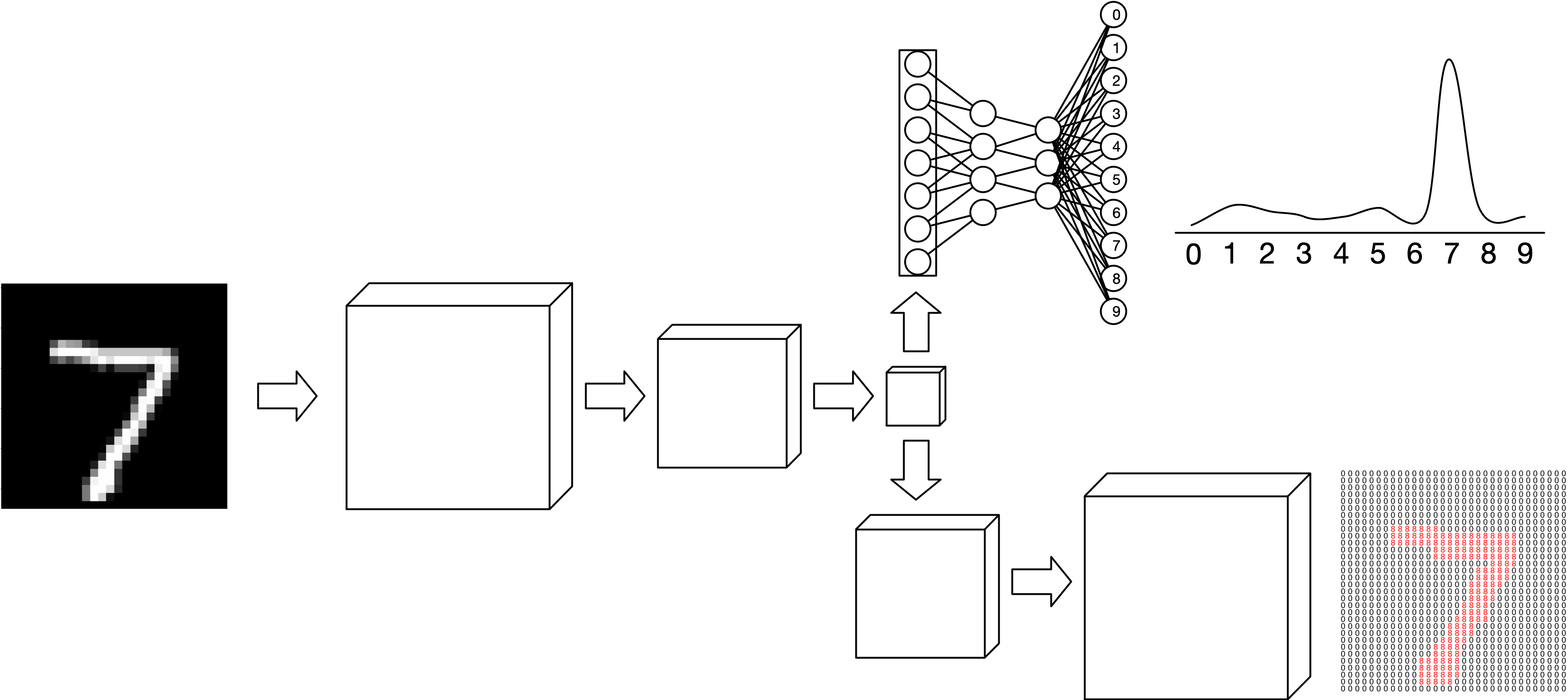}}
\end{center}
\caption{This figure illustrates the overall architecture of (a) main network schema for CvS in comparison to (b) standard vanilla classification network, and (c) a standard multi-task learning network.}
\label{fig:all_arch}
\end{figure*}
\textbf{Transfer learning.} With emerge of ImageNet, transfer learning have been commonly employed in many computer vision tasks by transferring the network weights learned on ImageNet classification. In general these method have shown to be an effective way of improving the performance of downstream tasks with small data by exploiting the weights of pretrained networks \cite{zhuang2020comprehensive}. Big Transfer \cite{kolesnikov2019big} is currently one of the strongest methods that harness the power of transfer learning to mitigate the problem of small dataset for classification tasks. However, transfer learning based methods require a large amount of data for building a rich base model. Moreover, transfer learning methods not only require careful selection of sets of information to transfer but also tend to perform to their full potential when the nature of data for the target task does not vary significantly from the data used in the pre-trained networks \cite{williams2020limits}. Additionally, the representation in some layers of the pretrained model may be local and irrelevant to the target task which can make the decision on number of layers to keep or remove less confident and more challenging \cite{yosinskiunderstanding}.

\textbf{Few-shot learning.} Few-shot learning methods are based on building reliable models that generalize to new classes with insufficient data in training set \cite{fei2006one, fink2004object}. In general, these methods require a large initial or base dataset that is of the same domain and it's the support set that can be small and hence cannot perform well when the initial dataset is also small. These methods often learn a similarity metric on training data and transfer it to new classes \cite{vinyals2016matching, snell2017prototypical, wu2018improving}. Unlike the few-shot learning methods our proposed model, CvS, aims to solve the classification tasks for extremely small datasets that do not have access to any initial large datasets to begin with, such as medical imaging classification problems. 

\textbf{Weakly supervised learning and label propagation.} Weakly supervised algorithms aim to build strong predictive model by learning from noisy or incomplete supervision where coarse labels or small subset of data with ground truth labels are available \cite{schwenker2014partially, hernandez2016weak, zhou2018brief}. Similarly, label propagation creates more training data by propagating from few labeled examples to a large collection of unlabeled examples \cite{xiaojin2002learning}. Deep Metric Transfer \cite{liu2019deep} improves upon label propagation by utilizing metric transfer to address the problem of object recognition from a very small amount of labeled data. This method, transfer a similarity metric learned from another related domain and propagate the labels from labeled examples to unlabeled images to enlarge the labeled data that enables training of deep neural networks. Although our proposed framework is similar to these methods, they do not fit our experimental set up for direct comparison. Weakly supervised learning often tries to show how the weak supervision is close to the full label supervision and thus, showing how label propagation is close to full manual labeling. However, our main goal with CvS is to show its superiority over classification, especially with respect to human annotation time. We acknowledge that although other semi-supervised methods could potentially improve upon the label propagation method, the focus of CvS is not on getting the actual label propagation to work but rather using the most naive and simple approach to show the lower bound of the performance.  
\begin{figure*}[htb]
	\centering
	\includegraphics[width=95mm]{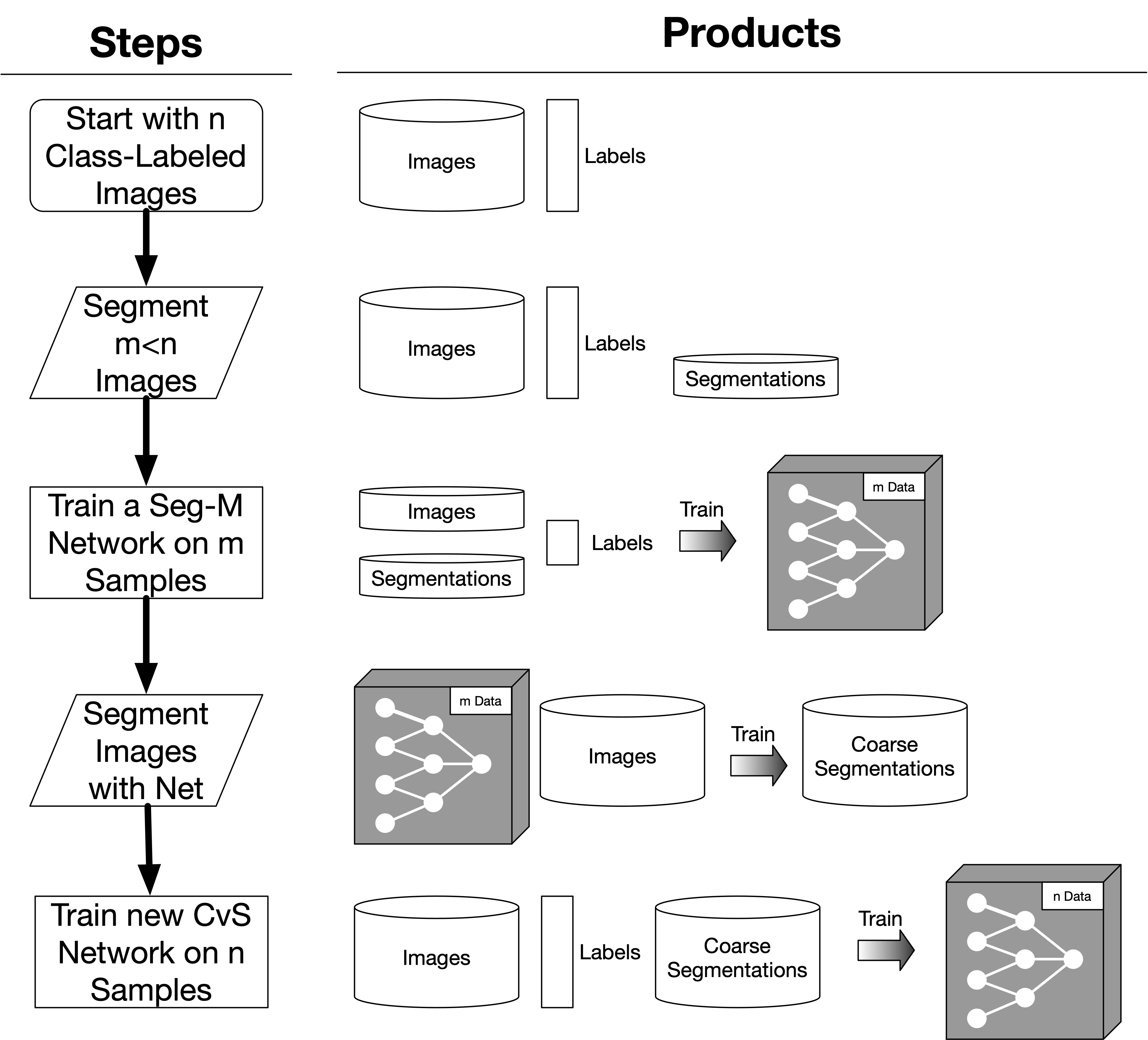}
	\caption{\textbf{Pipeline for Segmentation Propagation}}
	\label{fig:seg_propagate}
\end{figure*}

\section{Methodology}
CvS is a single-headed approach to the standard multi-task learning methods, which eliminates the need to balance the losses from different heads while still doing both tasks of segmentation and classification. CvS harnesses the power of segmentation to learn from smaller datasets, which allows us to perform classification on extremely small datasets ($\sim1-5$ samples per class). 

\subsection{Problem Formulation}
The CvS framework derives the classification label through a segmentation module. The overall architecture is illustrated in Fig.\ref{fig:cvs}

Suppose the training set comprises $N$ samples and denoted as 
\[\mathcal{D} = \big\{(x_k, s_k, y_k)|s_{k}(i,j)\in\{0,1\}, y_k \in\{1,2,...,P\} \big\}^{N}_{k=1}\]
where $x_k$ is of size $[H, W, C]$, $s_{k}$ represents the segmentation map, $y_k$ represents the ground truth class label, and $P$ represents the number of classes. Given $\mathcal{D}$, our goal is to learn a classifier $f_c:\cal{X}\xrightarrow{}\cal{Y}$ parameterized by $\theta_c$. We define functions $f(x_k;\theta_f)=z_k$ and $g(z_k;\theta_g)=h_k$ 
%
%
%
where $f(.)$ parameterized by $\theta_f$ represents the backbone function $f:\cal{X}\xrightarrow{}\cal{H}$ mapping the input image to the latent feature map $z_k$, and $g(.)$ parameterized by $\theta_g$ represents the head function $g:\cal{H}\xrightarrow{}\cal{S}$ mapping the feature encoding $z_k$ to the segmentation map, $h_k$, of size $[H, W, P+1]$. The extra class (class zero) in $p+1$ represents the background and is discarded. Then we define another function $q(h_k)=y_k$ that averages over the remaining $P$ segmentation maps followed by a softmax function to obtain the class label for the given input. Given the input $x_k \in \cal{X}$, function $f_c$ can be decomposed such that 
\begin{equation}\label{eq:decoder_clf}
f_c(x_k, \theta_c)=(q\circ g\circ f)(x_k)
\end{equation}
where $\theta_c=\{\theta_g, \theta_f\}$. The backbone function $f(.)$ is modeled by a neural network and the head network consists of a stack of convolutional blocks. The loss function is taken to be a cross entropy loss.

\subsection{CvS Segmentation}
Most classification datasets do not provide segmentation labels and collecting segmentation labels for the full classification dataset is prohibitively expensive. Motivated by this challenge, we employ two simple approaches, binarization and label propagation to procure segmented datasets. Although label propagation has been widely used, we employ it in the context of our work by learning a preliminary model from segmenting a small subset of the dataset ($\sim1-5$ samples per class) and use this model to propagate segmentation labels to the rest of the dataset. This allows us to apply CvS on the whole dataset of any size by collecting segmentation labels for an extremely small subset. 

\subsubsection{Binarization}
For datasets containing black and white images such as MNIST, we opted for simplicity and applied a binarization technique with threshold $0.0$ to obtain the segmentation maps. In the obtained image, the $0.0$ pixel values represent the background and the non-zero pixels (pixels with value $1$) represent the class that the given image belongs to. 

\subsubsection{Segmentation Label Propagation}
For datasets with more complex image data such as CIFAR10/100 where the binarization method was no longer applicable, we employed label propagation to obtain segmentation labels for the dataset. The overall pipeline for segmentation propagation is illustrated in Fig.\ref{fig:seg_propagate}. We started by manually segmenting $M$ samples per class (Usually chosen to be a very small number). Then we used the $M\times n_{classes}$ data points to train the segmentation network of the CvS framework (the functions $f$ and $g$) which we refer to as Seg-M network. Then we used this preliminary model to propagate the segmentation labels to the rest of the images.   

\section{Model Architecture}
The CvS framework classifies the given input image via a segmentation module which itself is composed of two main components, backbone and head. The backbone network takes the input image and learns the latent feature maps which are being utilized by the head network to predict segmentation maps. The predicted segmentation maps are further used to derive the class label for the input image. The overall architecture is illustrated in Fig.\ref{fig:cvs}. 

\subsection{Backbone Architecture}
The CvS framework allows various choices of backbone architecture without any constraints. We adopted ResNet-101 and Wide-ResNet for our work. We employed a custom Wide-ResNet \footnote{We borrowed the same network architecture developed by Bumsoo Kim https://github.com/meliketoy and replicated all the experiments for the purpose of our work.} with the depth and width set to $28$ and $10$ respectively. The network is composed of a convolutional layer followed by a stack of three ResNet blocks where each consists of two BatchNormalization-ReLU-Conv structures. The first layer in each ResNet block is followed by Dropout. For ResNet-101, we employed a standard architecture used in TorchVision package.

\subsection{Head Architecture}
The CvS head is built upon a convolutional layer. The architecture of the convolutional head varies slightly depending on the choice of the backbone network. When Wide-ResNet is used as the backbone, the head follows the BatchNormalization-ReLU-ConvTranspose structure. When ResNet101 is used, the DeepLabHead architecture is employed as the head layer. 

\section{Experiments}
\subsection{Data Collection}

\textbf{MNIST \cite{lecun2010mnist}:} MNIST dataset comprises $60K$ $28\times28$ hand written digits labeled with $10$ classes corresponding to digits $0$ to $9$. we employed the binarization method to obtain the segmentation labels. Given the $[0,1]$ valued images, zero-valued pixels were labeled as background, and pixels with value $1$ were multiplied by their corresponding class label, i.e. $1\times7$ representing the class of digit $7$. Further, to distinguish the class of digit zero and the background class (zero-valued pixels), we incremented the non-zero pixel values by one i.e. $1\times7+1$ representing the class of digit $7$.   

\textbf{CIFAR10/100 \cite{krizhevsky2009learning}:} The CIFAR-10 and CIFAR-100 dataset consist of $60K$ $32\times32$ color images with $10$ and $100$ classes respectively. To obtain the segmentation labels, we employed the label propagation method. We chose the CIFAR-10 dataset and manually segmented $M$ images per class where $M\in \{1, 5, 10, 25, 50, 100\}$. Then we trained the Seg-M model using the $M\times10$ images with segmentation labels. Out of the six trained Seg-M networks, we chose Seg-10 and Seg-100 as the preliminary models to segment the rest of the dataset. For CIFAR-100, we employed the same Seg-10 and Seg-100 models from CIFAR-10 to obtain segmentation labels for the whole CIFAR-100 dataset.

\begin{figure}[]
	\centering
	\includegraphics[width=65mm]{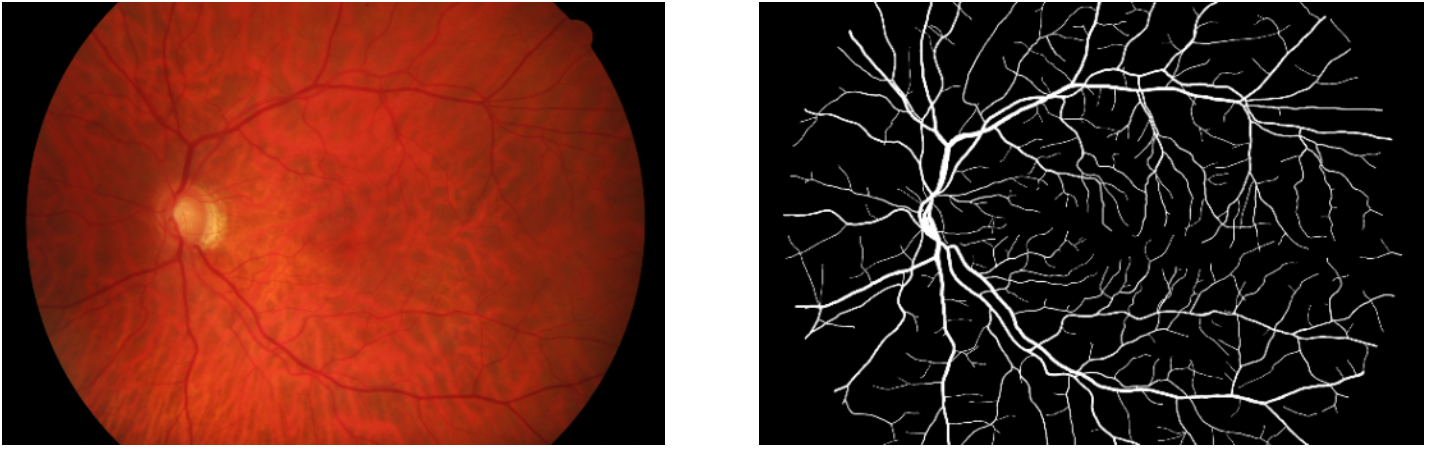}
	\caption{Sample of Fundus photo and its vessel segmentation from HRF.}
	\label{fig:fundus}
\end{figure}

\textbf{HRF \cite{budai2013robust}:} This dataset consists of $45$ Fundus images belonging to $45$ patients in three classes, healthy, diabetic retinopathy, or glaucomatous with $15$ images per class. This dataset provides the binary gold standard vessel segmentation for each image. A randomly selected sample of Fundus photo and its corresponding ground truth vessel segmentation is illustrated in Fig. \ref{fig:fundus}.

\subsection{Baselines}
To show the effectiveness of our proposed framework, we compared CvS with standard classifier deep neural networks, multi-task learning, and previous works for each image classification task. The baseline methods have a similar structure to CvS with ResNet101 or Wide-ResNet chosen as their backbone networks. 

\textbf{Linear Head:}
The linear head is employed in standard classification networks and follows the 
BatchNormalization-ReLU-AveragePooling structure followed by a linear layer.

\textbf{Multi-task Head:}
The multi-task learning method is composed of two segmentation and classification. When using Wide-ResNet as the backbone, we employed a ConvTranspose layer after applying BatchNormalization-ReLU for the segmentation head. For the classification head, we applied BatchNormalization-ReLU-AveragePooling followed by a linear layer. When using ResNet101 as the backbone network, we employed a stack of three ConvTranspose layers for segmentation head where the first two layers are followed by ReLu and BatchNorm. For classification, we applied an average pooling layer followed by a linear layer.
\begin{table*}[]
\small
\centering
\caption{MNIST Test Set Performance}
\label{tab:mnist-res}
\begin{tabular}{llllllllll}
\hline
\multicolumn{1}{c}{\multirow{2}{*}{Methods}} & \multicolumn{9}{c}{Number of samples per class} \\ \cline{2-10} 
\multicolumn{1}{c}{} & 1 & 5 & 10 & 25 & 50 & 100 & 500 & 1k & Full \\ \hline
{\ul \textit{ResNet101}} &  &  &  &  &  &  &  &  &  \\
Classification & 58.21 & 50.38 & 68.09 & 82.78 & 91.09 & 94.06 & 98.62 & 98.68 & 99.47 \\
Multi-task & 48.82 & 79.43 & 83.26 & 93.94 & 95.19 & 96.65 & \textbf{99.36} & 99.39 & \textbf{99.75} \\
CvS & \textbf{71} & 87.67 & \textbf{92.7} & 95.59 & \textbf{97.8} & 97.8 & 99.11 & 99.17 & 99.62 \\ \hline
{\ul \textit{W-ResNet}} &  &  &  &  &  &  &  &  &  \\
Classification & 20.16 & 77.05 & 78.78 & 84.88 & 89.31 & 95.45 & 98.49 & 98.83 & 99.16 \\
Multi-task & 15.25 & 30.87 & 37.17 & 84.6 & 96.28 & 97.87 & 99.18 & \textbf{99.51} & 99.45 \\
CvS & 54 & \textbf{88} & 90.9 & \textbf{95.68} & 97.41 & \textbf{97.95} & 99.08 & 99.25 & 99.51 \\ \hline
{\ul \textit{Others}} &  &  &  &  &  &  &  &  &  \\
LeNet & 47.7 & - & 72 & - & - & 82 & - & - & 98.5 \\ \hline
\end{tabular}
\end{table*}
\subsection{Experimental Setting}
We employed different settings based on the choice of the backbone network. When using Wide-ResNet as the backbone, the model is trained from random initialization, and images were kept in their original resolution for MNIST, CIFAR-10, and CIFAR-100. When ResNet101 is used, the pretrained network’s weights were used as initialization, and images were resized to $128\times128$ for MNIST, CIFAR-10, and CIFAR-100 and $512\times512$ for HRF.  

We used slightly different data augmentation based on the choice of dataset and the model. For CvS, when using MNIST, a random turn, random shift with zoom, and gaussian noise are employed. When using CIFAR-10, a random turn, color distortion, and random flip are used. When using CIFAR-100, a color distortion, random shift with zoom, and random flip are used. For baseline methods (classification and multi-task), when using CIFAR-10/100 and MNIST, a random crop and resize with random horizontal flip are employed. When using HRF, random horizontal flip and random turn are used. Depending on the size of the dataset, the batch size is selected from $\{8, 16, 32, 128\}$. For optimizer, we used SGD with momentum parameter set to $0.9$, weight decay $0.0005$, and initial learning rate $0.1$.  

\subsection{Experimental Results}
To show the effectiveness of our proposed method in a low data regime, we train our model using different data sizes ranging from only one sample per class to using the full dataset. 

\subsubsection{MNIST}
We selected M random samples per class of our training set where $M \in \{1, 5, 10, 25, 50, 100, 500, 1000, N\}$ and $N$ represents the size of the dataset. We compared our proposed method against a standard classification network, multi-task approach, and previous approach LeNet \cite{byerly2020branching}. The result is depicted in Table \ref{tab:mnist-res}. The result for LeNet is reported directly from their paper.  

The result in Table \ref{tab:mnist-res}, shows that CvS outperforms all the baseline methods significantly especially when the size of the dataset is very small ($M\leqslant100$) and achieves comparable results to multi-task learning when the number of training samples is larger than $50k$. We observed that using ResNet-101 as the backbone network achieves superior results over the Wide-ResNet when only a handful of examples per class ($M=1$) are available to the model and achieves comparable results in the larger data regime ($M\geqslant5$). The result indicates that CvS can achieve high performance without the need for a large amount of data for pre-training or a significant amount of computational resources.  

\subsubsection{CIFAR-10}\label{sec:cifar10exp}
Similar to the MNIST setting, we selected $M \in \{1, 5, 10, 25, 50, 100, 500, 1000, N\}$ random samples per class as our training data. Table \ref{tab:cifar10-res} compares the results of our work with the classification network and two of the previously proposed approaches Big Transfer \cite{kolesnikov2019big} and Deep Metric Transfer \cite{liu2019deep} where the results are reported directly from their papers. 
\begin{table*}[htb]
\small
\centering
\caption{Classification accuracy on CIFAR-10 test set given different numbers of samples per class and methodology.}
\label{tab:cifar10-res}
\begin{tabular}{llllllllll}
\hline
\multicolumn{1}{c}{\multirow{2}{*}{Methods}} & \multicolumn{9}{c}{Number of samples per class} \\ \cline{2-10} 
\multicolumn{1}{c}{} & 1 & 5 & 10 & 25 & 50 & 100 & 500 & 1k & Full \\ \hline
{\ul \textit{ResNet101}} &  &  &  &  &  &  &  &  &  \\
Classification & 18.85 & 20.67 & 26.01 & 27.12 & 37.37 & 42.16 & 69.75 & 76.4 & 93.55 \\
CvS & \textbf{39.31} & \textbf{67.24} & \textbf{73.94$^*$} & \textbf{80.95} & \textbf{86.43} & \textbf{90.1$^*$} & - & - & - \\
CvS(Seg-10) & - & - & - & 78.69 & 84.89 & 88.51 & 93.26 & 94.66 & 96.42 \\
CvS(Seg-100) & - & - & - & - & - & - & \textbf{93.79} & \textbf{95.37} & \textbf{97.13} \\ \hline
{\ul \textit{W-ResNet}} &  &  &  &  &  &  &  &  &  \\
Classification & 16.66 & 25.03 & 26.11 & 35.47 & 42.34 & 54.2 & 79.29 & 85.49 & 93.71 \\
CvS & 19.35 & 33.07 & 38.56$^{**}$ & 51.4 & 59.69 & 68.5$^{**}$ & - & - & - \\
CvS(Seg-10) & - & - & - & 45.45 & 54.09 & 62.12 & 78.93 & 84.81 & 93.29 \\
CvS(Seg-100) & - & - & - & - & - & - & 80.76 & 85.81 & 93.18 \\ \hline
{\ul \textit{Others}} &  &  &  &  &  &  &  &  &  \\
Big Transfer & \textbf{67} & \textbf{94} & \textbf{97} & - & - & \textbf{98} & - & - & \textbf{99.4} \\
DeepMetric-tr & - & 56.3 & 63.5 & - & 74.8 & 79.4 & 84.6 & 87.9 & - \\ \hline
\end{tabular}
\end{table*}

Table \ref{tab:cifar10-res} shows the result for CvS where $M \in \{1, 5, 10, 25, 50, 100\}$ samples per class were manually segmented. We then selected two of the trained CvS models achieving an accuracy of $73.94\%$ and $90.1\%$ with ResNet-101 and $38.56\%$ and $68.5\%$ with W-ResNet (depicted by $*$ and $**$ in Table \ref{tab:cifar10-res}) and chose their corresponding trained segmentation network, Seg-10 and Seg-100 as the preliminary models to segment the remaining images. The CvS(Seg-10) and CvS(Seg-100) in Table \ref{tab:cifar10-res} indicate the performance of CvS framework when Seg-10 and Seg-100 were employed to obtain the segmentation labels. 

As Table \ref{tab:cifar10-res} suggests, CvS outperforms standard classification networks and Deep Metric Transfer significantly. The performance gap is particularly noticeable in a low data regime. This result suggests that CvS is much more powerful as opposed to standard classifiers when dealing with extremely small datasets. We can also see that CvS achieves its best performance when using ResNet-101 as the backbone network.

As was expected, CvS models that had access to data with segmentation labels perform slightly better than those with predicted segmentation labels. Comparing the results of CvS(Seg-10) and CvS(Seg-100) shows that the model does not benefit much from increasing the number of data with segmentation labels. This result indicates the effectiveness of CvS in achieving high performance with having access to only a handful of data with segmentation labels.

Although CvS doesn't outperform Big Transfer, this method tend to perform to its full potential when the initial and target problems are related. Therefore its application becomes limited for domains such as medical imaging where the target data is significantly different from the ImageNet data \cite{williams2020limits, yosinskiunderstanding}.
\subsubsection{CIFAR-100}
For CIFAR-100 we selected $M \in \{1, 5, 10, 25, 50, 100, N\}$ random samples per class where $N$ is the size of the dataset. Since we did not segment any of the CIFAR-100 images, the segmentation annotation is performed by Seg-10 and Seg-100 networks from the CvS model trained on CIFAR-10 (depicted by $*$ in Table \ref{tab:cifar10-res}). Table \ref{tab:cifar100-res} compares the results of our work with the classification network and Big Transfer \cite{kolesnikov2019big}.

The observation from Table \ref{tab:cifar100-res} supports the results in Table \ref{tab:cifar10-res}. The result shows the superiority of CvS over standard classifier networks. We can also see that the model doesn't benefit from more data with segmentation labels. The overall result indicates that the CvS model can achieve much higher performance than the vanilla classification network with a negligible cost for manual segmentation.  

\subsubsection{High Resolution Fundus Photographs (HRF)}
In this section, we evaluate our work on a medical dataset for a real-world application of ophthalmic disease classification and compare it with standard classification networks, multi-task learning, and previous work from \cite{diaz2019cnns}. 

We employed 5-fold cross-validation on 45 images, where folds were chosen randomly and each fold used 36 images in the training set and was tested on the left-out 9 images. We used the hyperparameters that worked best for the CIFAR-10 experiments. Our baseline ResNet101 classification network achieved an accuracy of 66.67\%; our multitask network achieved an accuracy of 70.23\%, and our CvS network achieved an accuracy of 82.22\%. Our CvS method even slightly outperforms the high performance of 80\% reported by \cite{diaz2019cnns}, that uses a combination of $5$ datasets as their training data and does not use full Fundus but a crop around the optic disc to help increase signal to noise.
\begin{table*}[htb]
\small
\centering
\caption{Classification accuracy on CIFAR-100 test set given different numbers of samples per class and methodology. All segmentation labels were propagated from CIFAR10 trained networks, following the same nomenclature as Sec.\ref{sec:cifar10exp}.}
\label{tab:cifar100-res}
\begin{tabular}{lccccccc}
\hline
\multicolumn{1}{c}{\multirow{2}{*}{Methods}} & \multicolumn{7}{c}{Number of samples per class} \\ \cline{2-8} 
\multicolumn{1}{c}{} & 1 & 5 & 10 & 25 & 50 & 100 & Full \\ \hline
{\ul \textit{Backbone:ResNet101}} &  &  &  &  &  &  &  \\
Classification & 2.97 & 7.6 & 11.25 & 24.9 & 38.2 & 55.48 & 78.24 \\
CvS (Seg-10) & \textbf{21.78} & \textbf{46.93} & \textbf{56.4} & \textbf{65.21} & \textbf{70.24} & \textbf{75.14} & \textbf{83.9} \\
CvS (Seg-100) & 21.49 & 45.73 & 52.8 & 64.07 & 70.23 & 74.65 & 83.64 \\ \hline
{\ul \textit{Backbone:Wide-ResNet}} &  &  &  &  &  &  &  \\
Classification & 3.89 & 7.12 & 9.38 & 31.45 & 33.75 & 55.85 & 78.01 \\
CvS (Seg-10) & 10.23 & 19.35 & 24.54 & 33.51 & 41.89 & 53.02 & 75.76 \\
CvS (Seg-100) & 10.79 & 20.03 & 25.11 & 33.45 & 40.75 & 51.46 & 72 \\ \hline
{\ul \textit{Other Architectures}} &  &  &  &  &  &  &  \\
Big Transfer (SoTA) & \textbf{40} & \textbf{78} & \textbf{84} & \textbf{87} & - & \textbf{91} & \textbf{93.5} \\ \hline
\end{tabular}
\end{table*}
\begin{figure}[ht]
\begin{center}
\subfigure[CIFAR-10 Experiments]{\label{fig:cifar10_plot}\includegraphics[width=70mm]{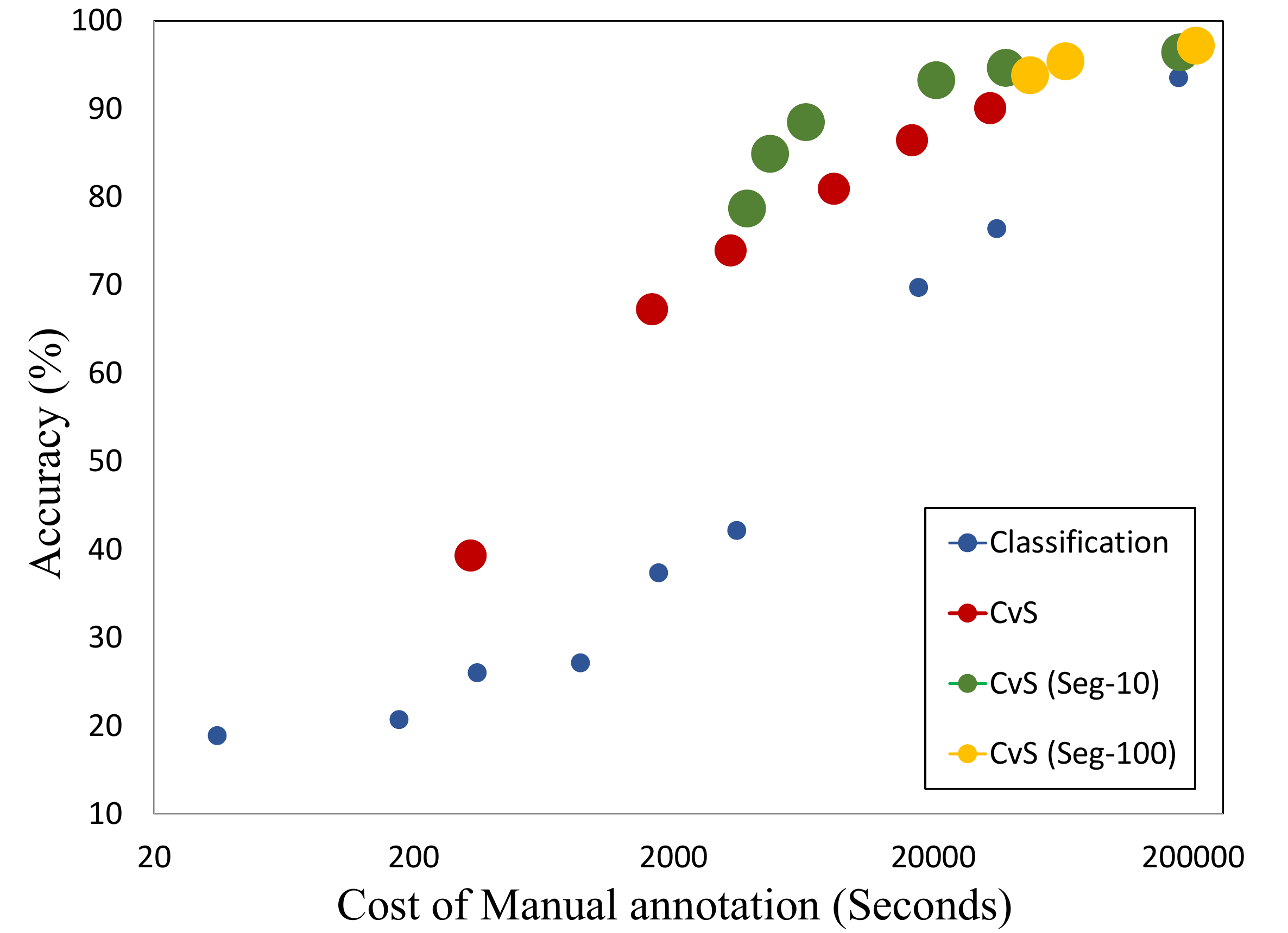}}
\end{center}
\begin{center}
\subfigure[CIFAR-100 Experiments ]{\label{fig:cifar100_plot}\includegraphics[width=70mm]{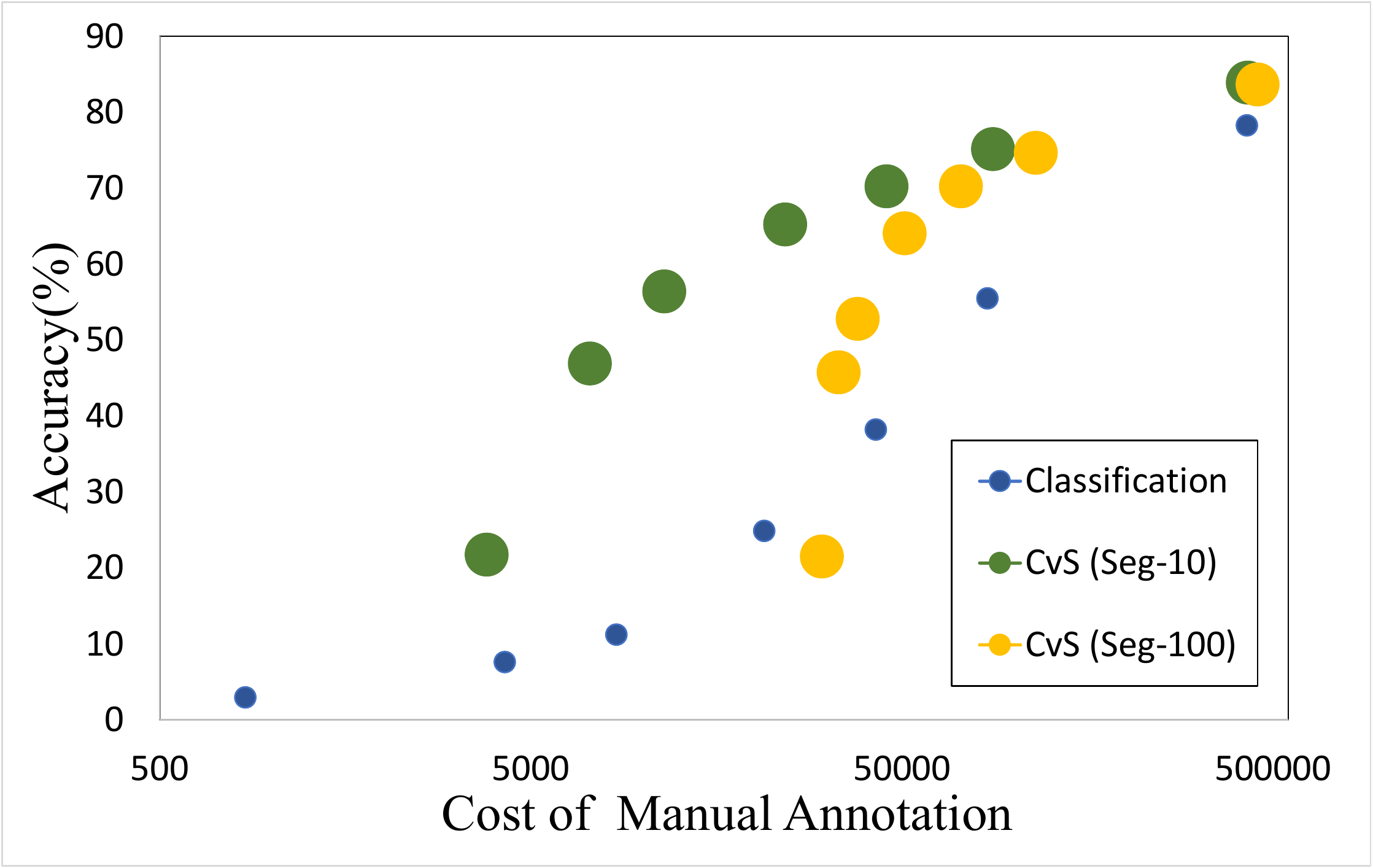}}
\end{center}
\caption{Comparison of the cost in terms of human annotation for different methods for CIFAR-10 and CIFAR-100. The radius of circles represent the computational costs of each method.}
\label{fig:cost_plot}
\end{figure}
\begin{figure*}[ht]
	\centering
	\includegraphics[width=95mm]{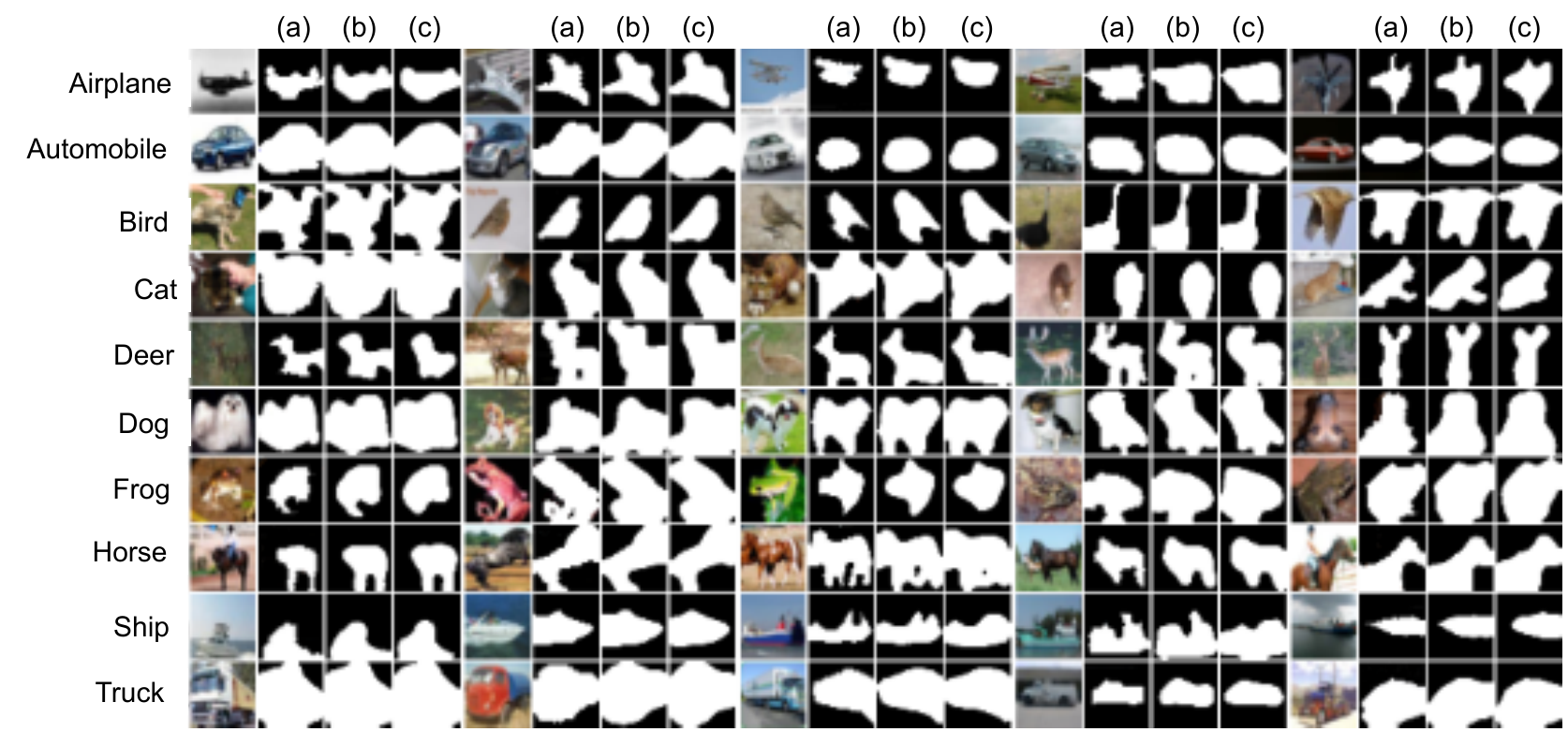}
	\caption{Illustration of predicted segmentation maps performed by label propagation technique for CIFAR-10. From left to right, we have an original image randomly selected from each class, (a) manual segmentation, (b), and (c) predicted result of Seg-10 and seg-100.}
	\label{fig:cifar10-images}
\end{figure*}

\begin{figure*}[ht]
	\centering
	\includegraphics[width=130mm]{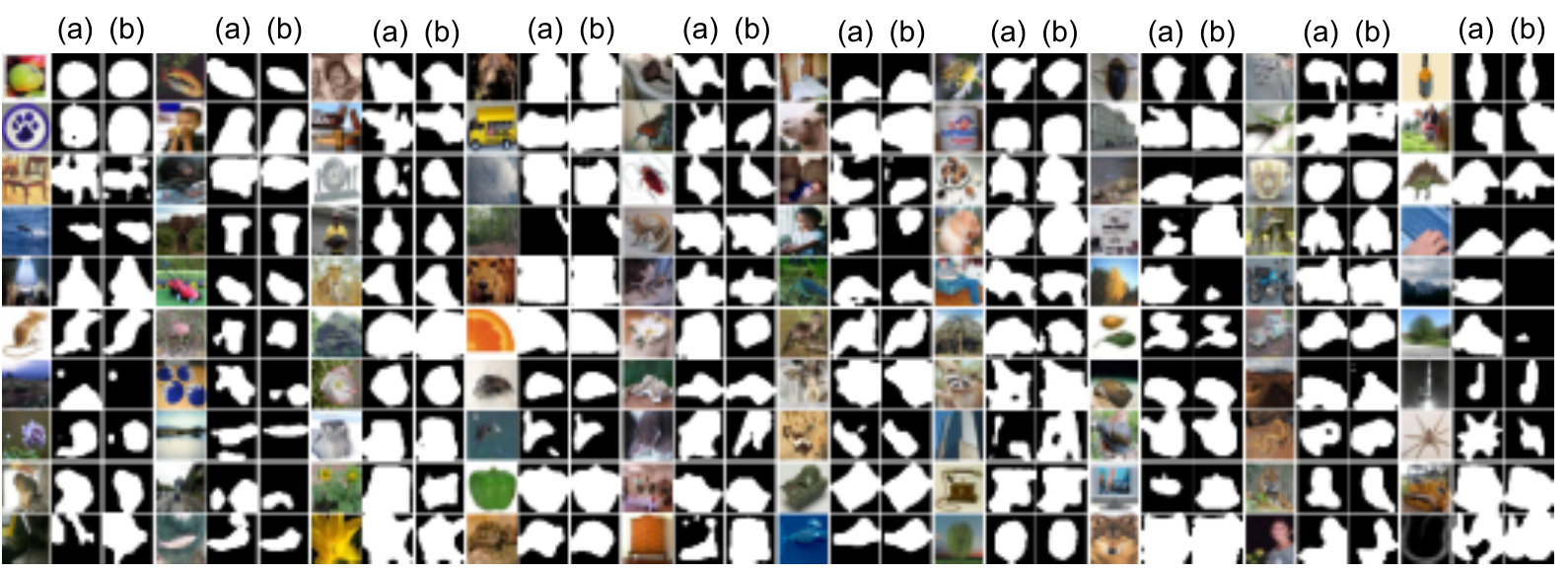}
	\caption{Illustration of predicted segmentation maps performed by (a) Seg-10 and (b) Seg-100 networks of CvS model for CIFAR-100.}
	\label{fig:cifar100-images}
\end{figure*}
\subsection{Cost Analysis}
In this section we evaluate the cost of the manual annotation for pure classification based approaches and our CvS model which is computed as follows: (1) the cost of pure classification is calculated as the average time spent for annotating each image with its corresponding class label, (2) the cost of CvS method is computed as the average time spent for manual segmentation of each image plus the time for manual annotation of each image with it corresponding class label. Figure \ref{fig:cost_plot} shows the cost analysis of compared methods for CIFAR-10 and CIFAR-100 from Table \ref{tab:cifar10-res} and Table \ref{tab:cifar100-res} using the ResNet-101 as the backbone. In our experiment, the average time for classification and segmentation annotation per example of CIFAR-10 was $3.5$ and $29.52$ seconds respectively. For CIFAR-100, the average time for classification annotation was $8.5$ second. As Fig.\ref{fig:cost_plot} suggests, all CvS plots are above the pure classification method. We can see that as the number of samples per class increases, the extra segmentation annotation of $1000$ more images doesn't matter relative to the $50k$ class labels at the upper limit. This shows that our CvS method is not only a better method for every dataset size, but also better given human annotation time.

\subsection{Label Propagation Analysis}
In this section, we analyze the quality of predicted segmentation labels performed by the segmentation propagation method. We employed our two preliminary models, Seg-10 and Seg-100, and visualized their results on a randomly selected image from each class in CIFAR-10 and 10 class selected from CIFAR-100. For CIFAR-10, we further compared the result against manually segmented images. The result for CIFAR-10 and CIFAR-100 are shown in Fig.\ref{fig:cifar10-images} and Fig.\ref{fig:cifar100-images} respectively. compares the result against images with segmentation labels. 

As Fig.\ref{fig:cifar10-images} and Fig.\ref{fig:cifar100-images} suggest, our proposed method shows a reasonable accuracy for predicted segmentation maps. The results suggest that the label propagation technique can obtain a quality segmented dataset with access to only a handful of data points with segmentation labels.

\section{Conclusion}
In this work, we presented a novel framework for classification in a low data regime. We studied its components and showed the effectiveness of our method on diverse classification tasks. Our experiments showed considerable improvement over previous approaches. Our method differs from multi-task learning in the choice of the loss function and tackles the issues of loss balancing from different heads in MTL by computing both tasks together. We also employed label propagation and binarization technique to alleviate the difficulty of procuring segmentation labels for classification datasets. 

\bibliographystyle{IEEEtran}
\bibliography{refs}

\end{document}